\documentclass[conference]{IEEEtran}
\IEEEoverridecommandlockouts

\usepackage{cite}
\usepackage{amsmath,amssymb,amsfonts}
\usepackage{algorithmic}
\usepackage{graphicx}
\usepackage{textcomp}

\usepackage[utf8]{inputenc} 
\usepackage[T1]{fontenc}    
\usepackage{hyperref}       
\usepackage{url}            
\usepackage{booktabs}       
\usepackage{amsfonts}       
\usepackage{nicefrac}       
\usepackage{microtype}      
\usepackage[table]{xcolor}
\definecolor{mygray}{gray}{0.85}
\usepackage{multirow}
\usepackage{algorithm}      
\usepackage{lipsum}
\usepackage{enumitem}
\setlist[itemize]{leftmargin=1.2em}
\usepackage{caption}
\usepackage{float} 
\usepackage{subfigure}
\usepackage{subcaption}
\usepackage{makecell}

\def\BibTeX{{\rm B\kern-.05em{\sc i\kern-.025em b}\kern-.08em
    T\kern-.1667em\lower.7ex\hbox{E}\kern-.125emX}}
\begin{document}

\title{HiFloat4 Format for Language Model Inference}

\author{
\IEEEauthorblockN{
Yuanyong Luo*, Jing Huang,Yu Cheng, Ziwei Yu, Kaihua Tang, Xinda Ma, Xin Wang, Anping Tong,\\
Guipeng Hu, Yun Xu, Mehran Taghian, Peng Wu, Guanglin Li, Yunke Peng, Tianchi Hu,\\ 
Minqi Chen, Michael Bi Mi, Hu Liu, Xiping Zhou, Junsong Wang, Qiang Lin, Heng Liao} 
\IEEEauthorblockA{\textit{Huawei}\\
*Email: luoyuanyong@\{hisilicon.com, yeah.net\}} 
}

\maketitle

\begin{abstract}

This paper introduces HiFloat4 (HiF4), a block floating-point data 
format tailored for deep learning. 
Each HiF4 unit packs 64 4-bit elements with 32 bits of shared scaling 
metadata, averaging 4.5 bits per value. 
The metadata specifies a three-level scaling hierarchy, 
capturing inter- and intra-group dynamic range while improving the 
utilization of the representational space. 
In addition, the large 64-element group size enables matrix 
multiplications to be executed in a highly fixed-point manner, 
significantly reducing hardware area and power consumption. 
To evaluate the proposed format, we conducted inference experiments 
on several language models, 
including LLaMA, Qwen, Mistral, DeepSeek-V3.1 and LongCat. 
Results show that HiF4 achieves higher average accuracy than the 
state-of-the-art NVFP4 format across multiple models and diverse 
downstream tasks.
\end{abstract}


\section{Introduction}

\begin{figure*}[htbp]
  \centering
  \includegraphics[width=17cm]{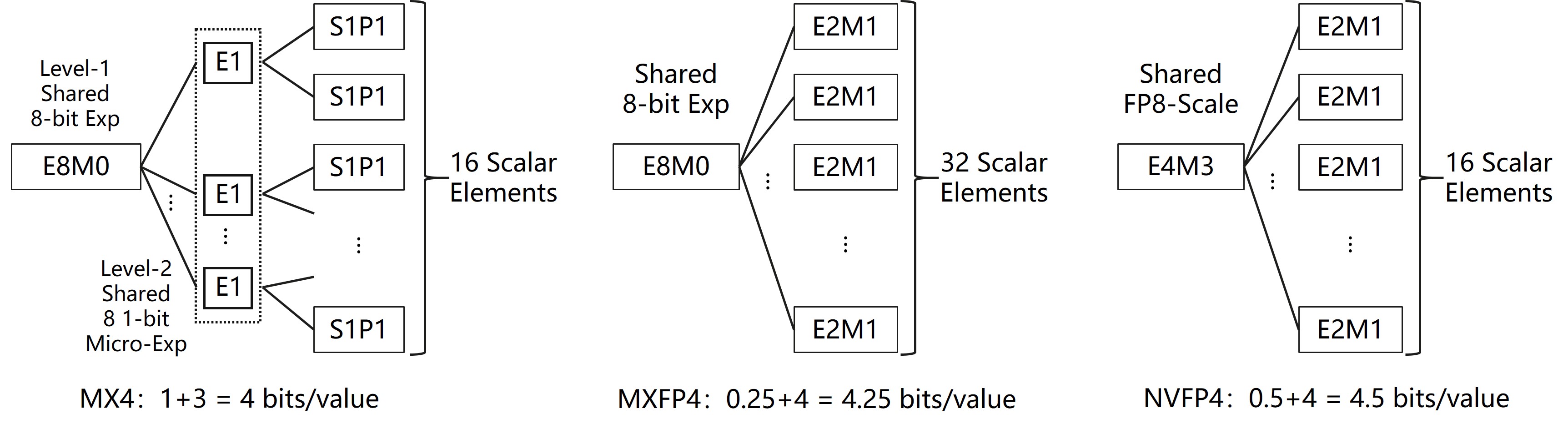}
  \caption{The Structure of Three 4-bit Block Floating-Point Formats}
  \label{BFP4-Structure}
\end{figure*}

In recent years, the rapid growth of large language models (LLMs) 
\cite{Naveed2025} has intensified 
the trade-offs among computational throughput, memory capacity, and energy 
efficiency \cite{Gholami2024}. 
These challenges have driven the community to investigate low-precision 
data formats as a promising solution. 
Several emerging 8-bit formats, including FP8 \cite{Nvidia2022}, 
HiF8 \cite{Eric2025, Luo2024}, and MXFP8 \cite{Nvidia2025, AMD2025}, 
have already been deployed in commercial hardware products to accelerate 
deep learning workloads. 
Yet, as model sizes and context lengths continue to expand, the exploration 
of low-precision formats has naturally advanced toward the 4-bit frontier. 
Although MXFP8, as a Block Floating-Point (BFP) format, 
trades off performance for training stability, 
BFP currently appears to be the most practical pathway for 4-bit 
representation. 
Because further reductions in bit-width can only be 
realized by more effectively exploiting exponent redundancy within 
localized data.  
In the following, we examine 3 influential 4-bit BFP designs outlined 
in Figure \ref{BFP4-Structure},  
each representing a technical advancement and introducing a new perspective.

\noindent\textbf{MX4: Shared Micro-Exponents} \\
To characterize intra-group variation, Microsoft and Meta proposed the MX4 
format \cite{DarvishRouhani2023}, 
which augments the shared 8-bit exponent \cite{DarvishRouhani2020} with 
8-way 1-bit micro-exponents, 
forming a two-level hierarchical scaling structure. 
However, since MX4 uses a group size of 16,  
the metadata incurs a relatively high overhead of 1 bit per value. 
To keep the average storage below 5 bits per element, 
the in-group sign-magnitude format must be reduced from 4 to 3 bits, 
resulting in an average storage cost of 4 bits per value. 
This reduction ultimately causes MX4 to deliver even lower accuracy than 
the vanilla 4-bit BFP format \cite{DarvishRouhani2020}.  
Consequently, AMD's Versal AI Edge series adopted only the related MX6 and 
MX9 formats, abandoning MX4 \cite{AMD2024}. 
In summary, MX4 illustrates the conceptual value of shared micro-exponents 
for capturing intra-group variation, 
yet the metadata cost — exacerbated by the small group size — prevents the 
format from achieving practical accuracy and efficiency. 

\noindent\textbf{MXFP4: Individual Micro-Exponent} \\ 
OCP-MXFP4 assigns an individual 2-bit micro-exponent to each 
element, enabling finer intra-group variation \cite{Rouhani2023a}. 
Typically, the 4-bit in-group format is the sign-magnitude S1P2, 
where S denotes the sign bit, the 1 before P indicates a 1-bit integer part, 
and the 2 after P indicates a 2-bit fractional part (P is the binary point).
This is equivalent to the common E1M2 in floating-point style, where the 
sign bit is usually omitted. 
E1M2 provides at most 3-bit significand but only a limited dynamic range of 
log2(1.75/0.25) = 2.81 binades.  
MXFP4 replaces E1M2 with E2M1, sacrificing 1 bit significand to introduce 
a 2-bit micro-exponent, thereby expanding the dynamic range to 
log2(6/0.5) = 3.58 binades. 
The group size is also enlarged to 32, with an average storage cost of 4.25 
bits per value. 
MXFP4 has been adopted by several hardware vendors, including 
Nvidia \cite{Micikevicius2025}, AMD \cite{AMD2025}, and Huawei \cite{Eric2025}. 
However, it is now widely recognized that MXFP4 suffers from significant 
accuracy degradation when applied to both activations and weights 
\cite{Rouhani2023}. 
As a result, it is currently employed only for weight-only quantization, 
as in OpenAI's GPT-OSS \cite{Agarwal2025}, leaving the substantial 4-bit 
computing power unused. 
In summary, MXFP4 introduces a new trade-off between significand precision 
and intra-group dynamic range, 
but still fails to achieve the full potential of 4-bit BFP. 

\noindent\textbf{NVFP4: Floating-Point Scale} \\
In addition to MXFP4, the Blackwell architecture also supports a proprietary 
NVFP4 format, 
which leverages the advantages of E2M1 over E1M2 and introduces two changes 
\cite{Micikevicius2025}. 
First, power-of-two scale cannot guarantee to normalize each group's peak 
magnitude to the representable upper bound of E2M1, 
wasting intra-group dynamic range. 
NVFP4 replaces it with a fine-grained FP8-E4M3 floating-point scale, 
fully utilizing E2M1's expressive space. 
Second, NVFP4 adopts the same 16-element group size 
as MXF4 \cite{DarvishRouhani2023}, 
raising the average storage cost from 4.25 to 4.5 bits per value compared to 
MXFP4, but effectively reducing outlier-driven quantization error. 
Consequently, NVFP4 achieves better accuracy than MXFP4 and can be used for 
both weights and activations, 
thereby initially unlocking practical 4-bit BFP computing power. 
The benefits, however, come at two heavy costs. 
First, the dynamic range of E4M3 is not wide enough. 
NVFP4 requires additional software-based per-tensor scaling (PTS) for both 
inference \cite{Alvarez2025} and training \cite{Abecassis2025}, 
incurring performance degradation during format conversion. 
Second, when considering higher computing power compared to 8-bit formats, 
pairing a floating-point scale with a small group size leads to substantial 
area and power overhead in matrix-compute units. 
In summary, NVFP4 employs a floating-point scale to fully utilize the 
intra-group dynamic range, 
albeit with substantial hardware and software cost. 

From a comprehensive standpoint, the preceding analysis underscores the 
inherent complexity of designing a 4-bit BFP format. 
MX4 and MXFP4 characterize intra-group variation using distinct 
micro-exponent mechanisms, while NVFP4 further emphasizes the importance 
of maximizing intra-group dynamic range utilization. 
Each approach entails trade-offs across multiple design dimensions, 
such as metadata overhead, group size, significand precision, 
and both intra- and inter-group dynamic ranges. 
Despite the applicability of NVFP4 to both weights and activations, 
its limitations in accuracy, inter-group dynamic range, and hardware 
efficiency remain pronounced. 
These shortcomings highlight the ongoing challenge of realizing practical 
and efficient 4-bit BFP computation.

Through multiple rounds of analysis and iteration, 
we identified a suitable technical entry point and proposed a new 4-bit BFP 
format, HiF4. 
Compared with the state-of-the-art, HiF4 delivers higher accuracy, broader 
inter-group dynamic range suitable for both inference and training, 
and lower hardware cost. 
In this paper, we describe the specification of HiF4, including the format 
definition and conversion. 
We then present comparisons of quantization error and dot-product compute 
flow between different 4-bit BFP formats. 
Finally, extensive experimental results on LLM inference are 
reported to demonstrate the feasibility and advantages of HiF4.

\section{HiFloat4}

This section provides a systematic description of the proposed 4-bit BFP 
format HiF4, and details its structural components. 
We then present the conversion algorithm from BF16 to HiF4, 
along with the suggested hardware support. 

\subsection{Format Definition}
\begin{figure}[htbp]
  \centering
  \includegraphics[width=\linewidth]{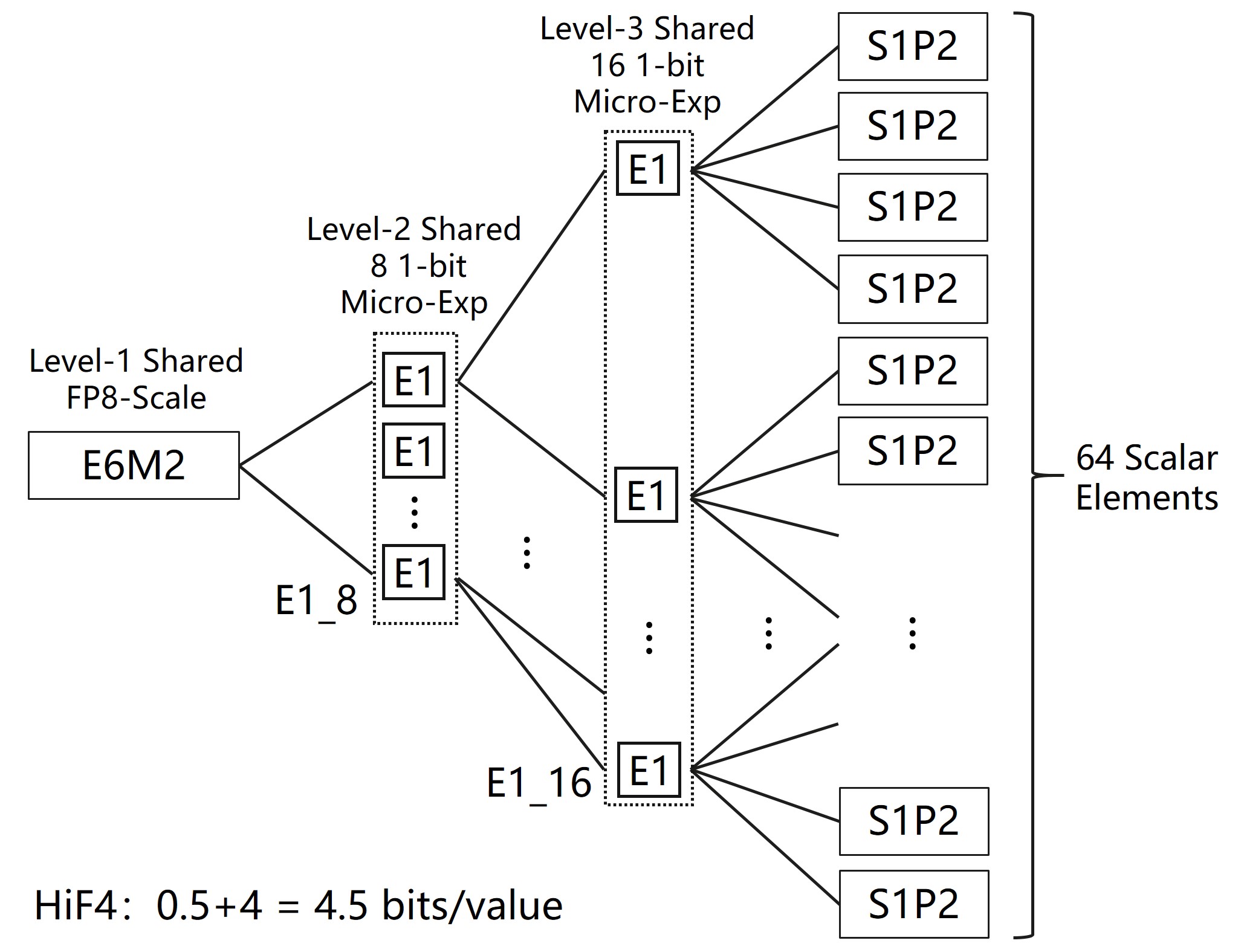}
  \caption{The Structure of HiF4 Block Floating-Point Format}
  \label{HiF4-Structure}
\end{figure}

As shown in Figure \ref{HiF4-Structure}, a basic HiF4 unit consists of 
32 bits shared scaling metadata, and 64 4-bit in-group elements, 
resulting in an average storage of 4.5 bits per value. 
The following defines each component in detail. 

\subsubsection{Scaling Metadata}

The metadata defines a three-level scaling hierarchy. 
The first level (E6M2) is a specially designed unsigned 8-bit 
floating-point format. 
This scale provides a wide dynamic range across groups and normalizes 
each group's peak magnitude to the representable upper bound of the 
remaining hierarchical format structure, 
ensuring full utilization of the intra-group dynamic range. 
Table \ref{E6M2-S1P2} summarizes the encoding details of E6M2. 
We assign 6 bits to the exponent field with a bias of 48, and 2 bits 
to the mantissa field with one hidden integer bit set to 1. 
E6M2 encodes NaN (Not a Number) but no other special values such 
as infinity or zero. 
Meanwhile, only normal mode is supported. 
Thus if we denote the unbiased exponent as E and the mantissa 
as M, an E6M2 number used in HiF4 should be interpreted as follows: 
\begin{equation}
  X = 2^{E} \times 1.M       \label{E6M2-Eq}
\end{equation} 
The second (E1\_8) and third (E1\_16) levels, implemented as 8-way 
and 16-way 1-bit micro-exponents, expand the intra-group dynamic 
range by refining local exponent differences, 
effectively mitigating the impact of outliers and suppressing 
quantization error. 
Micro-exponent E1 directly encodes 1 and 0, representing a very  
fine-grained power-of-two scaling factor. 
In this hierarchical scaling scheme, 
the level-1 E6M2 factor provides a global base scale connected to 
all level-2 micro-exponents. 
Each level-2 E1 further connects to two adjacent third-level 
micro-exponents, 
while each level-3 E1 in turn connects to four contiguous 4-bit 
elements within its local group. 
All three-level scaling metadata occupies 32 bits, 
distributed over 64 in-group elements, 
leading to an extra overhead of 0.5 bits per value. 

\begin{table}[tbp]
  \caption{E6M2 and S1P2 Encoding Details}
  \label{E6M2-S1P2}
  \centering
  \begin{tabular}{lll}
    \toprule 
     & Unsigned FP8-E6M2 & Sign-Magnitude S1P2\\
    \midrule
    Exponent Bias     & 48                                   & N/A    \\
    Unbiased Exp      & [-48, 15]                            & N/A    \\
    Infinity          & N/A                                  & N/A    \\
    Zero              & N/A                                  & S$0.00_2 = \pm 0.00$ \\
    NaN               & $111111\_ 11_2$                      & N/A    \\
    Max Value         & $111111\_10_2 = 2^{15} \times 1.50$  & S$1.11_2 = \pm 1.75$ \\
    Min Value         & $000000\_00_2 = 2^{-48} \times 1.00$ & S$0.01_2 = \pm 0.25$ \\
    \bottomrule
  \end{tabular}
\end{table}

\subsubsection{Element Encoding}

HiF4 encodes the 4-bit in-group elements using the 
sign-magnitude S1P2 representation. 
In this SXPY notation, S denotes the sign bit and P indicates the 
binary point. 
The value X preceding P designates a X-bit integer part, whereas 
the value Y following P designates a Y-bit fractional part. 
The encoding details of S1P2 are summarized in Table \ref{E6M2-S1P2}. 
Conceptually, S1P2 is equivalent to the E1M2 format in floating-point 
representation; 
however, we adopt the S1P2 notation because it is more intuitive to 
interpret.

\subsubsection{Represented Values}

Since we have elaborated on each component, 
we can now look at the HiF4 unit as a whole and 
compute its represented values. 
As depicted in Figure \ref{HiF4-Structure}, 
a single group in the HiF4 format comprises four 
levels of data, 
which are related through multiplication. 
Let $\{S1P2\}_i$ denote the i-th in-group element 
value, where i $\in$ [1, 64].
Let $\{E1\_8\}_j$ denote the j-th micro-exponent in the 
level-2 scaling metadata, where j $\in$ [1, 8]. 
Similarly, let $\{E1\_16\}_k$ denote the k-th 
micro-exponent in the level-3 scaling metadata, 
where k $\in$ [1, 16]. 
Finally, let $\{V_i\}_{i=1}^{64}$ represent the 64 real 
numbers in a HiF4 group. 
Based on these definitions, each value of a HiF4 unit 
can be expressed as follows: 
\begin{itemize}
  \item If $E6M2 = NaN$, then $V_i = NaN$ for all i $\in$ [1, 64] 
  \item Otherwise, 
  \begin{equation}
    V_i = E6M2 \times {\color{blue} 2^{(\{E1\_8\}_{\lceil i/8 \rceil} + 
      \{E1\_16\}_{\lceil i/4 \rceil})} \times \{S1P2\}_i} 
      \label{HiF4 Equation}
  \end{equation}
\end{itemize}
  
As formulated in Equation \ref{HiF4 Equation}, 
the remaining blue part — beyond the global base scale E6M2 — 
indicates the intra-group dynamic range that a HiF4 unit can 
cover: 
the maximum positive value is $2^{(1+1)} \times 1.75 = 7$, 
while the minimum positive value is $2^{(0+0)} \times 0.25 = 0.25$. 
Thus the available intra-group dynamic range of HiF4 is 
log2(7/0.25) = 4.81 binades. 
Some typical values and features of HiF4 and the state-of-the-art 
NVFP4 \cite{Abecassis2025} are outlined in Table \ref{HiF4-NVFP4}.

\begin{table}[tbp]
  \caption{Typical Values and Features for HiF4 and NVFP4}
  \label{HiF4-NVFP4}
  \centering
  \begin{tabular}{lll}
    \toprule 
     & HiF4 & NVFP4 \\
    \midrule
    Storage Overhead      & 4.5 bits/value & 4.5 bits/value \\
    Group Size            & 64             & 16             \\
    Special Values        & NaN and $\pm 0$& NaN and $\pm 0$\\
    4-bit Element         & S1P2 (E1M2)    & E2M1           \\
    Significand Precision & 3 bits         & 2 bits         \\
    Global Base Scale     & E6M2           & E4M3           \\
    Max Positive Value    & $2^{18}\times1.3125$ & $2^{11}\times1.3125$ \\
    Min Positive Value    & $2^{-50}$ & $2^{-10}$ \\
    Global Dynamic Range  & [-50, 18]: 69 binades  & [-10, 11]: 22 binades \\
    Local Dynamic Range   & 4.81 binades   & 3.58 binades   \\
    \bottomrule
  \end{tabular}
\end{table}

\subsection{Format Conversion} \label{HiF4_convert}

Algorithm \ref{BF16 to HiF4} illustrates the computational flow for 
converting a vector of 64 BF16 data into a HiF4 unit. 
The similar conversion procedure applies to other high-precision formats, 
such as FP32 and FP16.

\begin{algorithm}
  \caption{Conversion from BF16 to HiF4}
  \label{BF16 to HiF4}
  \begin{algorithmic}[1]
    \REQUIRE{A 64-length BF16 Vector: $V64$} 
    \ENSURE{A HiF4 Unit: E6M2, E1\_8, E1\_16, S1P2\_64} 
    \FOR{$i$ = 1 to 16}
      \STATE 16 Local Peak Magnitudes: \\
      $V16[i] = \max(|V64[4 \times i - 3 : 4 \times i]|)$ 
    \ENDFOR 
    \FOR{$i$ = 1 to 8}
      \STATE 8 Local Peak Magnitudes: \\
      $V8[i] = \max(V16[2 \times i - 1 : 2 \times i])$ 
    \ENDFOR 
    \STATE Global Peak Magnitude: $Vmax = \max(V8)$
    \STATE High-Precision Scale Factor: \\
    $SF_{BF16} = Vmax \times (1/7)_{BF16}$
    \STATE Level-1 Scale Factor: \\
    $E6M2 = BF16\_to\_E6M2(SF_{BF16})$
    \STATE Reciprocal of E6M2: \\
    $E6M2\_REC = E6M2\_REC\_to\_BF16(E6M2)$
    \STATE Level-2 Scale Factors: \\
    $E1\_8 = (V8 \times E6M2\_REC \ge 4)?\  1:0 $
    \FOR{$i$ = 1 to 16}
      \STATE Level-3 Scale Factors: $E1\_16[i] = $ \\
      $(V16[i] \times E6M2\_REC \times 2^{-E1\_8[\lceil i/2 \rceil]} \ge 2)?\  1:0 $
    \ENDFOR
    \FOR{$i$ = 1 to 64}
      \STATE High-Precision Elements: $V64\_scaled[i] = $ \\
      $V64[i] \times E6M2\_REC \times 2^{-E1\_8[\lceil i/8 \rceil]} \times 2^{-E1\_16[\lceil i/4 \rceil]}$
    \ENDFOR
    \STATE In-Group S1P2 Elements: \\
    $S1P2\_64 = BF16\_to\_S1P2(V64\_scaled)$
  \end{algorithmic}  
\end{algorithm}

The format conversion algorithm is decomposed into three 
sequential stages. \\
\textbf{Stage 1} (lines 1-7) performs a three-level tree reduction 
on the 64 BF16 inputs. 
\begin{itemize}
  \item In the first level, adjacent groups of 4 elements are 
  compared in parallel, yielding 16 local maximum absolute values. 
  \item The second level pairwise-reduces 16 maxima into 
  8 values. 
  \item The final level obtains the global maximum absolute value 
  across the 8 results.
\end{itemize}
\textbf{Stage 2} (lines 8-14) derives the three-level hierarchical 
scaling metadata.
\begin{itemize}
  \item Lines 8-9 compute the level-1 scale factor, 
  in which 7 is the maximum absolute value that the intra-group 
  structure can represent, 
  and division by a constant is replaced by multiplication with 
  its reciprocal to avoid performance degradation.  
  Then a dedicated BF16 to E6M2 instruction is required to quantize  
  the scale into E6M2 format. 
  \item Lines 10-11 generate 8 level-2 micro-exponents, 
  in which a bespoke reciprocal instruction that operates directly 
  on E6M2 and returns BF16 eliminates explicit division. 
  Then 8 parallel multiply-compare operations produce the E1\_8. 
  \item Lines 12-14 evaluate 16 level-3 micro-exponents. 
  Observe that $2^{-E1}$ collapses to either 0.5 or 1. 
  Therefore, special bypass mode of multiply instruction can be 
  realized to compute the $BF16\times2^{-E1}$ value. 
\end{itemize}
\textbf{Stage 3} (lines 15-18) calculates the 64 in-group elements.  
\begin{itemize}
  \item Line 16 applies the three-level HiF4 scaling factors to the 
  original 64 BF16 inputs.
  \item Line 18 quantizes the scaled results into S1P2 format. 
  If the results after rounding exceed the representable range 
  of S1P2, it should be clamped to the nearest representable bound, 
  preserving the sign. 
\end{itemize}

All rounding operations in BF16 to HiF4 conversion should use 
round-half-to-even or round-half-away-from-zero. 
Moreover, since E6M2 has no subnormal numbers, 
the proposed E6M2\_REC\_to\_BF16 conversion instruction can 
be efficiently implemented using only a 4-entry lookup table 
indexed by the input 2-bit mantissa and simple exponent 
subtractions, 
to derive the output mantissa and exponent respectively. 
Meanwhile, some fused instructions such as multiply-compare 
and multiply-convert, can further accelerate the HiF4 
conversion process.

\section{Quantization Error and Dot Product} \label{Quantization Error and Dot Product}
This section first characterizes the quantization errors 
of different 4-bit BFP formats, 
then analyzes the complexity of dot product computation 
for HiF4 and NVFP4. 

\begin{figure}[tbp]
  \centering
  \includegraphics[width=\linewidth]{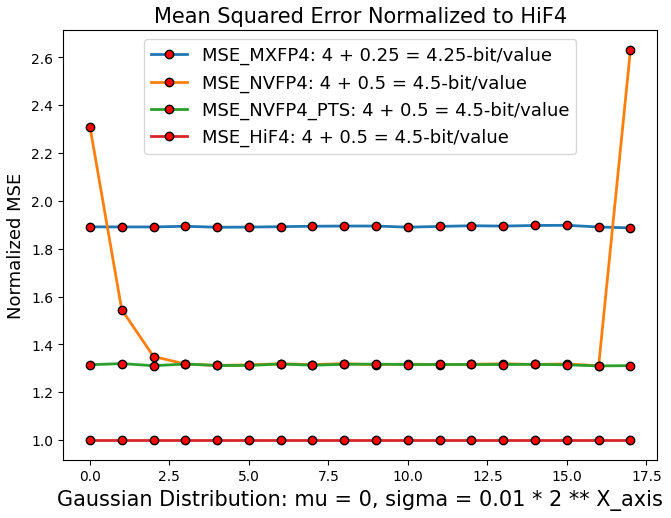}
  \caption{Quantization Error Comparison of 4-bit BFP Formats}
  \label{MSE-BFP4}
\end{figure}

\subsection{Quantization Error Evaluation}
We generated 18 high-precision Gaussian-distributed random 
matrices of size $1024 \times 1024$, 
each with zero mean and standard deviation $\sigma = 0.01 
\times 2^x$, where $x \in [0, 17]$. 
Every matrix was converted to three 4-bit BFP formats. 
HiF4 followed Algorithm \ref{BF16 to HiF4}. 
MXFP4 adopted the method described in \cite{Rouhani2023}. 
NVFP4 employed E4M3 scaling to normalize the peak magnitude 
of each group to 6, the representable upper bound of E2M1. 
Since only NVFP4 lacks sufficient dynamic range and requires 
additional software-based PTS, 
two versions were prepared for fair comparison: 
a direct cast and a pipeline that first scales each tensor's 
peak magnitude to 2688 and then quantizes to NVFP4 \cite{Alvarez2025}. 
We computed the mean squared error (MSE) relative to the 
original high-precision matrix for each format 
and normalized the results with respect to HiF4. 

Figure \ref{MSE-BFP4} depicts the quantization error 
of HiF4, MXFP4, and NVFP4. 
When data values approach the numeric bounds of NVFP4, 
overflow and underflow cause the quantization error to 
increase significantly. 
Applying PTS before format conversion can eliminate this 
precision loss. 
In contrast, HiF4 and MXFP4 do not require PTS, thus  
incurring no additional quantization overhead. 
Excluding NVFP4's fluctuation, the MSE ratio across the 
three 4-bit BFP formats remains stable: 
\[
\text{HiF4 : NVFP4 : MXFP4} = 1 : 1.32 : 1.89 
\]
Note that although the Gaussian distribution matches the 
centralized nature of the data in neural networks, 
the quantization error may vary in practical scenarios. 
The experimental results serve as a quantitative reference 
for analyzing the accuracy of different 4-bit BFP formats. 

\subsection{Dot Product Evaluation}

\begin{figure*}[htbp]
  \centering
  \includegraphics[width=17cm]{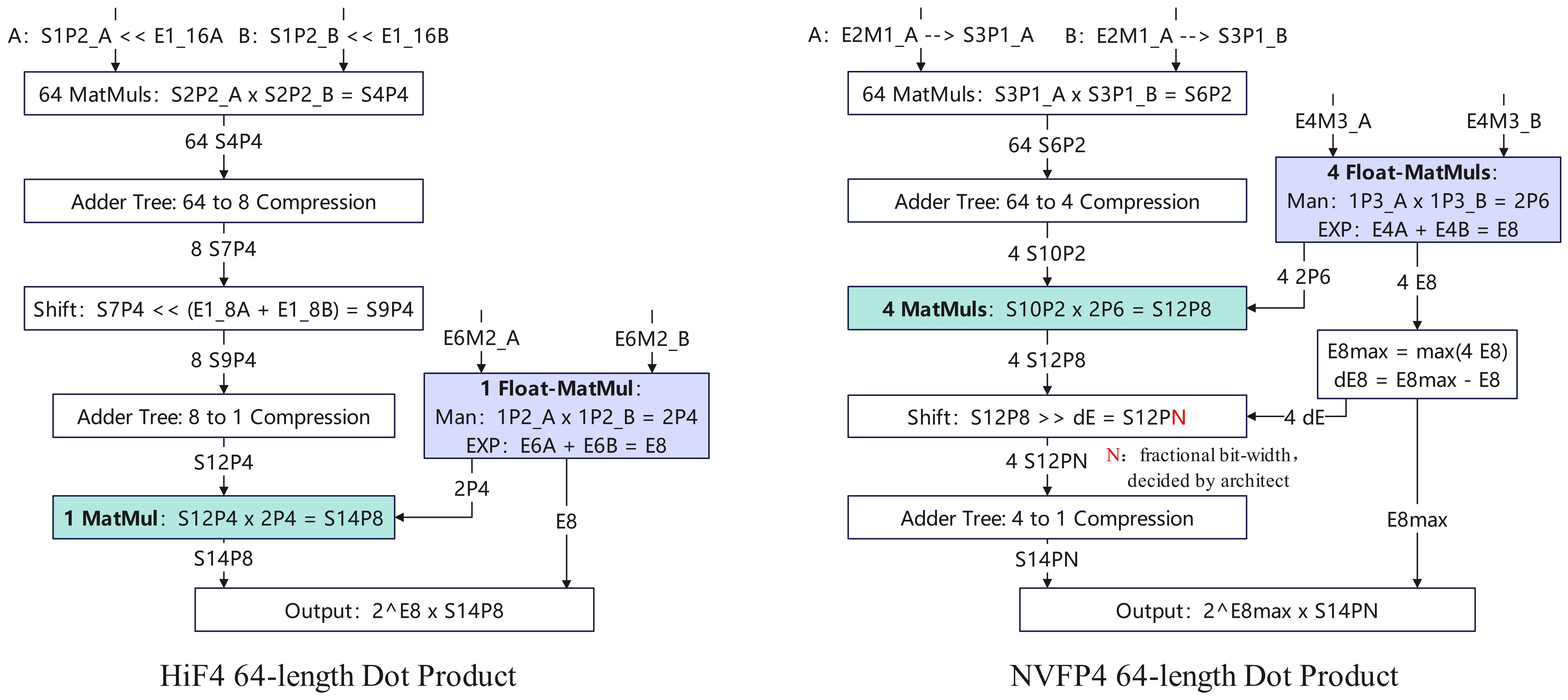}
  \caption{Compute Flow of 64-length Dot Product for HiF4 and NVFP4}
  \label{BFP4-MatMul}
\end{figure*}

Let A: 
$\{E6M2^{(A)}, E1\_8^{(A)}, E1\_16^{(A)}, S1P2\_64^{(A)}\}$ 
and B: $\{E6M2^{(B)}, E1\_8^{(B)}, E1\_16^{(B)}, S1P2\_64^{(B)}\}$ 
denote the two HiF4 units. 
The dot product of A and B should be computed as follows: 
\begin{align}
  Dot(A, B)& = E6M2^{(A)} \times E6M2^{(B)} \times \notag \\
  &\sum_{i=1}^8  2^{(E1\_8^{(A)}[i] + E1\_8^{(B)}[i])} \times \notag \\
  &\sum_{j = 2i-1}^{2i} 2^{(E1\_16^{(A)}[j] + E1\_16^{(B)}[j])} \times \notag \\
  &\sum_{k = 4j-3}^{4j} (S1P2\_64^{(A)}[k] \times S1P2\_64^{(B)}[k])
  \label{Dot-Product}
\end{align}
In hardware, a processing element (PE) that performs a K-length 
dot product and accumulation in parallel is a fundamental building 
block of matrix-compute units, 
such as Nvidia GPU's Tensor Cores \cite{Nvidia2017} and 
Ascend NPU's Cube Cores \cite{Liao2019}.
To double the computing power of 4-bit BFP formats relative to 
8-bit floating-point formats, 
both Tensor Cores and Cube Cores now require PEs to support 
64-length dot products \cite{Nvidia2025, Eric2025}.
Under these conditions, a single pair of HiF4 units suffices to 
match the PE input width, 
since their group size is exactly 64. 
In contrast, four pairs of NVFP4 units are required to support 
a 64-length dot product in a PE, 
because their group size is only 16.
Next, we briefly analyze the compute flow of HiF4 and NVFP4 in 
a 64-length dot product to reveal their differences in power and 
area consumption.

For the HiF4 dot product, level-2 and level-3 micro-exponents 
can be implemented as simple left-shift operations in hardware. 
There are many opportunities to absorb the micro-exponents into 
the computations related to the S1P2 elements, 
depending on the implementation strategy. 
In our example compute flow, we directly absorb the level-3 
micro-exponents into the S1P2 elements prior to multiplications. 
In this way, multiplier inputs become 5-bit integers, 
which can be denoted as S2P2 (for simplicity, we continue using 
the SXPY notation). 
Similarly, E2M1 elements in NVFP4 are converted into 
5-bit integers (S3P1) before multiplications. 
Figure \ref{BFP4-MatMul} illustrates the compute flow of 
64-length dot product for both formats. 
During the compression from 64 products to a single S12P4 output, 
HiF4 employs pure integer arithmetic, 
requiring only one small floating-point multiplier and one large 
integer multiplier at the final stage of the accumulation tree. 
In contrast, NVFP4 retains integer operations only during the 
reduction from 64 products to four S10P2 outputs, after which four  
small floating-point multipliers and four large integer multipliers 
are required. 
The final accumulation from four partial results to the output is 
then performed entirely in floating-point. 
Consequently, HiF4 eliminates six multipliers and simplifies the 
accumulation process compared with NVFP4. 

In practice, 4-bit BFP formats are integrated into 
existing dot-product units originally optimized for 16-bit 
(FP16 and BF16) and 8-bit (INT8 and Float8) formats. 
Although resource sharing is feasible across different precision 
modes, high-precision formats do not require intermediate and final 
multipliers within the accumulation tree. 
Thus, the additional multipliers introduced by floating-point 
metadata E6M2 and E4M3 incur extra hardware overhead. 
As a result, for 64-length dot product, HiF4 occupies only 
approximately one-third the incremental area of NVFP4 and reduces 
the power consumption by about 10\%. 
In summary, HiF4 enables a more area- and power-efficient 
implementation for matrix multiplication.

\begin{table*}[htbp]
\caption{Experimental Results of 4 Small LLMs Evaluated across 8 Benchmarks}
\label{tab3.2.1_llama}
\centering
\begin{tabular}{c | c | cccccccc | c} 
\hline
\hline

\rule{0pt}{2.2ex} Model & A-W Quant Type & ARC-C & ARC-E & BoolQ & HellaS & LamOp & Piqa & WinoG & MMLU & Mean \\

\hline 

\rule{0pt}{2.2ex} \multirow{10}{*}{\rotatebox{90}{Llama2-7B}} &  BF16 & 45.65 & 74.41 & 77.74 & 75.99 & 73.67 & 79.11 & 69.06 & 46.52 & 67.77 \\

\cline{2-11}
\rule{0pt}{2.2ex} & NVFP4 & 44.54 & 73.32 & 76.02 & 75.02 & 73.69 & 77.37 & 70.48 & 41.51 & 66.49\\

& \cellcolor{mygray}{\color{gray}\qquad{\small --- Acc Drop}} & \cellcolor{mygray}{\color{red}-1.11} & \cellcolor{mygray}{\color{red}-1.09} & \cellcolor{mygray}{\color{red}-1.72} & \cellcolor{mygray}{\color{red}-0.97} & \cellcolor{mygray}{\color{blue}+0.02} & \cellcolor{mygray}{\color{red}-1.74} & \cellcolor{mygray}{\color{blue}+1.42} & \cellcolor{mygray}{\color{red}-5.01} & \cellcolor{mygray}{\color{red}-1.28} \\

\rule{0pt}{2.2ex} & NVFP4+PTS & 44.62 & 72.01 & 76.67 & 74.92 & 73.16 & 77.91 & 66.69 & 43.43 & 66.18 \\

& \cellcolor{mygray}{\color{gray}\qquad{\small --- Acc Drop}} & \cellcolor{mygray}{\color{red}-1.03} & \cellcolor{mygray}{\color{red}-2.40} & \cellcolor{mygray}{\color{red}-1.07} & \cellcolor{mygray}{\color{red}-1.07} & \cellcolor{mygray}{\color{red}-0.51} & \cellcolor{mygray}{\color{red}-1.20} & \cellcolor{mygray}{\color{red}-2.37} & \cellcolor{mygray}{\color{red}-3.09} & \cellcolor{mygray}{\color{red}-1.59} \\

\rule{0pt}{2.2ex} & HiF4 & 44.63 & 74.08 & 76.73 & 75.00 & 72.65 & 78.40 & 68.39 & 44.54 & 66.80 \\

& \cellcolor{mygray}{\color{gray}\qquad{\small --- Acc Drop}} & \cellcolor{mygray}{\color{red}-1.02} & \cellcolor{mygray}{\color{red}-0.33} & \cellcolor{mygray}{\color{red}-1.01} & \cellcolor{mygray}{\color{red}-0.99} & \cellcolor{mygray}{\color{red}-1.02} & \cellcolor{mygray}{\color{red}-0.71} & \cellcolor{mygray}{\color{red}-0.67} & \cellcolor{mygray}{\color{red}-1.98} & \cellcolor{mygray}{\color{red}-0.97}\\

\rule{0pt}{2.2ex} & HiF4+HiGPTQ & 45.31 & 73.61 & 76.58 & 74.91 & 73.66 & 77.86 & 68.99 & 43.89 & 66.85 \\

& \cellcolor{mygray}{\color{gray}\qquad{\small --- Acc Drop}} & \cellcolor{mygray}{\color{red}-0.34} & \cellcolor{mygray}{\color{red}-0.80} & \cellcolor{mygray}{\color{red}-1.16} & \cellcolor{mygray}{\color{red}-1.08} & \cellcolor{mygray}{\color{red}-0.01} & \cellcolor{mygray}{\color{red}-1.25} & \cellcolor{mygray}{\color{red}-0.07} & \cellcolor{mygray}{\color{red}-2.63} & \cellcolor{mygray}{\color{red}-0.92}\\

\hline 

\rule{0pt}{2.2ex} \multirow{10}{*}{\rotatebox{90}{LLama3-8B}} & BF16 & 53.41 & 77.78 & 81.16 & 79.15 & 75.65 & 80.85 & 72.93 & 66.55 & 73.44 \\

\cline{2-11}
\rule{0pt}{2.2ex} & NVFP4 & 48.04 & 73.74 & 78.35 & 76.56 & 74.21 & 78.56 & 71.43 & 61.64 & 70.32 \\

& \cellcolor{mygray}{\color{gray}\qquad{\small --- Acc Drop}} & \cellcolor{mygray}{\color{red}-5.37} & \cellcolor{mygray}{\color{red}-4.04} & \cellcolor{mygray}{\color{red}-2.81} & \cellcolor{mygray}{\color{red}-2.59} & \cellcolor{mygray}{\color{red}-1.44} & \cellcolor{mygray}{\color{red}-2.29} & \cellcolor{mygray}{\color{red}-1.50} & \cellcolor{mygray}{\color{red}-4.91} & \cellcolor{mygray}{\color{red}-3.12} \\

\rule{0pt}{2.2ex} & NVFP4+PTS & 49.74 & 74.54 & 77.46 & 77.60 & 74.85 & 79.76 & 70.80 & 62.81 & 70.95\\

& \cellcolor{mygray}{\color{gray}\qquad{\small --- Acc Drop}} & \cellcolor{mygray}{\color{red}-3.67} & \cellcolor{mygray}{\color{red}-3.24} & \cellcolor{mygray}{\color{red}-3.70} & \cellcolor{mygray}{\color{red}-1.55} & \cellcolor{mygray}{\color{red}-0.80} & \cellcolor{mygray}{\color{red}-1.09} & \cellcolor{mygray}{\color{red}-2.13} & \cellcolor{mygray}{\color{red}-3.74} & \cellcolor{mygray}{\color{red}-2.49} \\

\rule{0pt}{2.2ex} & HiF4 & 51.15 & 76.73 & 79.85 & 77.84 & 73.76 & 79.09 & 71.79 & 63.37 & 71.70 \\

& \cellcolor{mygray}{\color{gray}\qquad{\small --- Acc Drop}} &  \cellcolor{mygray}{\color{red}-2.26} & \cellcolor{mygray}{\color{red}-1.05} & \cellcolor{mygray}{\color{red}-1.31} & \cellcolor{mygray}{\color{red}-1.31} & \cellcolor{mygray}{\color{red}-1.89} & \cellcolor{mygray}{\color{red}-1.76} & \cellcolor{mygray}{\color{red}-1.14} & \cellcolor{mygray}{\color{red}-3.18} & \cellcolor{mygray}{\color{red}-1.74}\\

\rule{0pt}{2.2ex} & HiF4+HiGPTQ & 52.18 & 77.46 & 80.40 & 77.47 & 72.85 & 79.60 & 71.87 & 63.55 & 71.92 \\

& \cellcolor{mygray}{\color{gray}\qquad{\small --- Acc Drop}} &  \cellcolor{mygray}{\color{red}-1.23} & \cellcolor{mygray}{\color{red}-0.32} & \cellcolor{mygray}{\color{red}-0.76} & \cellcolor{mygray}{\color{red}-1.68} & \cellcolor{mygray}{\color{red}-2.80} & \cellcolor{mygray}{\color{red}-1.25} & \cellcolor{mygray}{\color{red}-1.06} & \cellcolor{mygray}{\color{red}-3.00} & \cellcolor{mygray}{\color{red}-1.52}\\

\hline 

\rule{0pt}{2.2ex} \multirow{10}{*}{\rotatebox{90}{Qwen2.5-14B}} &  BF16 & 58.96 & 79.34 & 85.54 & 82.94 & 74.31 & 81.88 & 74.74 & 80.17 & 77.24\\

\cline{2-11}
\rule{0pt}{2.2ex} & NVFP4 & 58.53 & 80.56 & 83.49 & 81.38 & 72.95 & 81.72 & 74.43 & 76.53 & 76.20\\

& \cellcolor{mygray}{\color{gray}\qquad{\small --- Acc Drop}} & \cellcolor{mygray}{\color{red}-0.43} & \cellcolor{mygray}{\color{blue}+1.22} & \cellcolor{mygray}{\color{red}-2.05} & \cellcolor{mygray}{\color{red}-1.56} & \cellcolor{mygray}{\color{red}-1.36} & \cellcolor{mygray}{\color{red}-0.16} & \cellcolor{mygray}{\color{red}-0.31} & \cellcolor{mygray}{\color{red}-3.64} & \cellcolor{mygray}{\color{red}-1.04}\\

\rule{0pt}{2.2ex} & NVFP4+PTS & 56.57 & 80.18 & 85.72 & 81.45 & 72.87 & 81.66 & 73.16 & 78.63 & 76.28\\

& \cellcolor{mygray}{\color{gray}\qquad{\small --- Acc Drop}} & \cellcolor{mygray}{\color{red}-2.39} & \cellcolor{mygray}{\color{blue}+0.84} & \cellcolor{mygray}{\color{blue}+0.18} & \cellcolor{mygray}{\color{red}-1.49} & \cellcolor{mygray}{\color{red}-1.44} & \cellcolor{mygray}{\color{red}-0.22} & \cellcolor{mygray}{\color{red}-1.58} & \cellcolor{mygray}{\color{red}-1.54} & \cellcolor{mygray}{\color{red}-0.96}\\

\rule{0pt}{2.2ex} & HiF4 & 58.66 & 81.55 & 85.86 & 81.45 & 72.74 & 81.29 & 73.72 & 78.62 & 76.74 \\

& \cellcolor{mygray}{\color{gray}\qquad{\small --- Acc Drop}} & \cellcolor{mygray}{\color{red}-0.30} & \cellcolor{mygray}{\color{blue}+2.21} & \cellcolor{mygray}{\color{blue}+0.32} & \cellcolor{mygray}{\color{red}-1.49} & \cellcolor{mygray}{\color{red}-1.57} & \cellcolor{mygray}{\color{red}-0.59} & \cellcolor{mygray}{\color{red}-1.02} & \cellcolor{mygray}{\color{red}-1.55} & \cellcolor{mygray}{\color{red}-0.50}\\

\rule{0pt}{2.2ex} & HiF4+HiGPTQ & 60.71 & 83.04 & 86.27 & 81.38 & 73.02 & 81.67 & 75.06 & 78.65 & 77.48 \\

& \cellcolor{mygray}{\color{gray}\qquad{\small --- Acc Drop}} & \cellcolor{mygray}{\color{blue}+1.75} & \cellcolor{mygray}{\color{blue}+3.70} & \cellcolor{mygray}{\color{blue}+0.73} & \cellcolor{mygray}{\color{red}-1.56} & \cellcolor{mygray}{\color{red}-1.29} & \cellcolor{mygray}{\color{red}-0.21} & \cellcolor{mygray}{\color{blue}+0.32} & \cellcolor{mygray}{\color{red}-1.52} & \cellcolor{mygray}{\color{blue}+0.24}\\

\hline 

\rule{0pt}{2.2ex} \multirow{10}{*}{\rotatebox{90}{Mistral-7B}} &  BF16 &  52.39 & 78.37 & 82.17 & 80.50 & 75.14 & 82.21 & 74.11 & 63.30 & 73.52\\

\cline{2-11}
\rule{0pt}{2.2ex} & NVFP4 & 28.41 & 26.30 & 56.09 & 26.69 & 0.0 & 48.69 & 48.30 & 26.79 & 32.66\\

& \cellcolor{mygray}{\color{gray}\qquad{\small --- Acc Drop}} & \cellcolor{mygray}{\color{red}-23.98} & \cellcolor{mygray}{\color{red}-52.07} & \cellcolor{mygray}{\color{red}-26.08} & \cellcolor{mygray}{\color{red}-53.81} & \cellcolor{mygray}{\color{red}-75.14} & \cellcolor{mygray}{\color{red}-33.52} & \cellcolor{mygray}{\color{red}-25.81} & \cellcolor{mygray}{\color{red}-36.51} & \cellcolor{mygray}{\color{red}-40.87} \\

\rule{0pt}{2.2ex} & NVFP4+PTS & 50.77 & 77.19 & 80.86 & 79.71 & 73.96 & 80.85 & 72.45 & 61.15 & 72.12 \\

& \cellcolor{mygray}{\color{gray}\qquad{\small --- Acc Drop}} & \cellcolor{mygray}{\color{red}-1.62} & \cellcolor{mygray}{\color{red}-1.18} & \cellcolor{mygray}{\color{red}-1.31} & \cellcolor{mygray}{\color{red}-0.79} & \cellcolor{mygray}{\color{red}-1.18} & \cellcolor{mygray}{\color{red}-1.36} & \cellcolor{mygray}{\color{red}-1.66} & \cellcolor{mygray}{\color{red}-2.15} & \cellcolor{mygray}{\color{red}-1.41}  \\

\rule{0pt}{2.2ex} & HiF4 & 51.41 & 76.85 & 81.35 & 79.41 & 74.11 & 81.01 & 72.06 & 61.63 & 72.23 \\

& \cellcolor{mygray}{\color{gray}\qquad{\small --- Acc Drop}} & \cellcolor{mygray}{\color{red}-0.98} & \cellcolor{mygray}{\color{red}-1.52} & \cellcolor{mygray}{\color{red}-0.82} & \cellcolor{mygray}{\color{red}-1.09} & \cellcolor{mygray}{\color{red}-1.03} & \cellcolor{mygray}{\color{red}-1.20} & \cellcolor{mygray}{\color{red}-2.05} & \cellcolor{mygray}{\color{red}-1.67} & \cellcolor{mygray}{\color{red}-1.29}\\

\rule{0pt}{2.2ex} & HiF4+HiGPTQ & 51.84 & 77.74 & 82.22 & 79.48 & 73.31 & 81.91 & 73.40 & 61.53 & 72.68 \\

& \cellcolor{mygray}{\color{gray}\qquad{\small --- Acc Drop}} & \cellcolor{mygray}{\color{red}-0.55} & \cellcolor{mygray}{\color{red}-0.63} & \cellcolor{mygray}{\color{blue}+0.05} & \cellcolor{mygray}{\color{red}-1.02} & \cellcolor{mygray}{\color{red}-1.83} & \cellcolor{mygray}{\color{red}-0.30} & \cellcolor{mygray}{\color{red}-0.71} & \cellcolor{mygray}{\color{red}-1.77} & \cellcolor{mygray}{\color{red}-0.84}\\

\hline
\hline
\end{tabular}
\end{table*}

\section{Language Model Inference with HiFloat4}

This section conducts a comprehensive inference evaluation of 
Post-Training Quantization (PTQ) across diverse tasks and LLM 
architectures on HiF4 and NVFP4 formats.  

\subsection{Post-Training Quantization for LLMs}

To demonstrate the inherent precision of the 4-bit formats, 
we prioritize using HiF4 direct-cast, NVFP4 direct-cast, 
and the NVFP4 + PTS approach for LLM inference, as described 
in Section~\ref{Quantization Error and Dot Product}. 
Meanwhile, to show that many existing PTQ methods can be 
adapted to enhance HiF4 accuracy with minor changes, 
we developed a tailored PTQ method based on the vanilla 
GPTQ~\cite{Frantar2023}, termed HiGPTQ, 
to exploit the fine-grained structure of the HiF4 BFP format. 

\subsection{Experiments on Small LLMs}
We begin by evaluating our method on relatively smaller LLMs 
with the parameter size ranging from 7B to 14B. 
Detailed results and the corresponding inference accuracy degradation, 
denoted as Acc Drop, across eight benchmarks are presented in 
Table~\ref{tab3.2.1_llama}. 
To provide an overall view of quantization errors, 
Table~\ref{tab3.2.3_average} reports the average inference accuracy 
across all evaluated models (w/ and w/o the crashed Mistral-7B) for 
NVFP4, NVFP4+PTS, HiF4, and HiF4+HiGPTQ. 

\noindent\textbf{Models:} 
We evaluate 4-bit quantization accuracy across a diverse 
collection of LLM architectures and model scales, 
including LLaMA2-7B~\cite{Touvron2023}, 
LLaMA3-8B~\cite{Dubey2024}, 
Qwen2.5-14B~\cite{Qwen2025}, 
and Mistral-7B~\cite{Jiang2023}. 
These models cover a broad range of architectural designs, 
including Multi-Head Attention (MHA)~\cite{Vaswani2017} 
and Grouped-Query Attention (GQA)~\cite{Ainslie2023}, 
as well as multiple forms of Feed-Forward Networks 
(FFNs)~\cite{Shazeer2020}.

\begin{table}[htbp]
\caption{Average Inference Accuracy for Small LLMs}
\label{tab3.2.3_average}
\centering
\begin{tabular}{c|c|cccc}
    
\hline
\hline

\rule{0pt}{2.2ex} \# LLM Models     & BF16  & NVFP4              & \makecell{NVFP4+\\PTS} & HiF4               & \makecell{HiF4+\\HiGPTQ} \\
\hline 

\rule{0pt}{2.2ex} 4 (w/ Mistral-7B) & 72.99 & 61.42              & 71.38                  & \textbf{71.87}     & 72.23 \\
\rule{0pt}{2.2ex} — Acc Drop        & -     & \color{red}{Crash} & \color{red}{-1.61}     & \color{red}{-1.12} & \color{red}{-0.76} \\
\hline

\rule{0pt}{2.2ex} 3(w/o Mistral-7B) & 72.82 & 71.01              & 71.14                  & \textbf{71.75}     & 72.08 \\
\rule{0pt}{2.2ex} — Acc Drop        & -     & \color{red}{-1.81} & \color{red}{-1.68}     & \color{red}{-1.07} & \color{red}{-0.74} \\
\hline

\hline
\end{tabular}
\end{table}

\begin{table*}[htbp]
\caption{Experimental Results of DeepSeek-V3.1 and LongCat Evaluated across 10 Benchmarks}
\label{tab3.2.1_deepseek}
\centering

\begin{tabular}{c | c | cccccccccc | c} 
\hline
\hline

\rule{0pt}{2.2ex} Model & A-W Quant Type & ARC-C & ARC-E & BoolQ & HellaS  & Piqa & WinoG &Gsm8K & MMLU  & Math500  & CMMLU & Mean \\

\hline 

\rule{0pt}{2.2ex} \multirow{7}{*}{\rotatebox{90}{\makecell{DeepSeek-V3.1\\671B}}} &  BF16 & 79.91 & 84.44 & 79.76 & 84.41 & 92.93 & 89.34 & 94.46 &  84.86 &75.00 & 89.28
    & 85.44 \\

\cline{2-13}
\rule{0pt}{2.2ex} & NVFP4 & 82.83 & 86.85 & 76.91 & 81.18 & 91.29 & 87.92 & 95.00 & 84.31 &74.00&87.76& 84.81\\

& \cellcolor{mygray}{\color{gray}\qquad{\small --- Acc Drop}} & \cellcolor{mygray}{\color{blue}+2.92} & \cellcolor{mygray}{\color{blue}+2.41
} & \cellcolor{mygray}{\color{red}-2.85} & \cellcolor{mygray}{\color{red}-3.23} & \cellcolor{mygray}{\color{red}-1.64} & \cellcolor{mygray}{\color{red}-1.42} & \cellcolor{mygray}{\color{blue}+0.54} & \cellcolor{mygray}{\color{red}-0.55} & \cellcolor{mygray}{\color{red}-1.00}& \cellcolor{mygray}{\color{red}-1.52} & \cellcolor{mygray}{\color{red}-0.63} \\

\rule{0pt}{2.2ex} & NVFP4+PTS & 81.55 & 87.06 & 77.83 & 81.00 & 91.29 & 88.00 & 95.30 & 
84.72 & 78.20 & 87.44
    & 85.24\\

& \cellcolor{mygray}{\color{gray}\qquad{\small --- Acc Drop}} & \cellcolor{mygray}{\color{blue}+1.64} & \cellcolor{mygray}{\color{blue}+2.62} & \cellcolor{mygray}{\color{red}-1.93} & \cellcolor{mygray}{\color{red}-3.41} & \cellcolor{mygray}{\color{red}-1.64} & \cellcolor{mygray}{\color{red}-1.34} & \cellcolor{mygray}{\color{blue}+0.84} & \cellcolor{mygray}{\color{red}-0.14} & \cellcolor{mygray}{\color{blue}+3.20} & \cellcolor{mygray}{\color{red}-1.84} & \cellcolor{mygray}{\color{red}-0.20} \\

\rule{0pt}{2.2ex} & HiF4 & 80.86 & 86.89& 78.87 & 86.60 & 92.44 & 89.11 & 95.75 & 84.77 & 80.20  & 88.71 & 86.42\\

& \cellcolor{mygray}{\color{gray}\qquad{\small --- Acc Drop}} & \cellcolor{mygray}{\color{blue}+0.95} & \cellcolor{mygray}{\color{blue}+2.45} & \cellcolor{mygray}{\color{red}-0.89} & \cellcolor{mygray}{\color{blue}+2.19} & \cellcolor{mygray}{\color{red}-0.49} & \cellcolor{mygray}{\color{red}-0.23} & \cellcolor{mygray}{\color{blue}+1.29} & \cellcolor{mygray}{\color{red}-0.09} & \cellcolor{mygray}{\color{blue}+5.20} & \cellcolor{mygray}{\color{red}-0.57} & \cellcolor{mygray}{\color{blue}+0.98}\\

\hline 
\hline

\hline 

\rule{0pt}{2.2ex} \multirow{7}{*}{\rotatebox{90}{\makecell{LongCat\\560B}}} &  BF16 & 84.38 & 86.64 & 66.85 & 82.09 & 91.46 & 80.27 & 95.91 & 59.19&84.80&81.65&81.32 \\

\cline{2-13}
\rule{0pt}{2.2ex} & NVFP4 &84.72&88.50&62.78&84.42&89.39&79.24&96.29&38.81&78.60&72.12&77.49
\\

& \cellcolor{mygray}{\color{gray}\qquad{\small --- Acc Drop}} & \cellcolor{mygray}{\color{blue}+0.34}  & \cellcolor{mygray}{\color{blue}+1.86}& \cellcolor{mygray}{\color{red}-4.07
} & \cellcolor{mygray}{\color{blue}+2.33} & \cellcolor{mygray}{\color{red}-2.07} & \cellcolor{mygray}{\color{red}-1.03} & \cellcolor{mygray}{\color{blue}+0.38} & \cellcolor{mygray}{\color{red}-20.38} & \cellcolor{mygray}{\color{red}-6.20}& \cellcolor{mygray}{\color{red}-9.53} & \cellcolor{mygray}{\color{red}-3.84} \\

\rule{0pt}{2.2ex} & NVFP4+PTS &87.04&89.26&63.55&84.17 &89.83&79.40&95.45&39.00&78.20&72.24 &77.81\\

& \cellcolor{mygray}{\color{gray}\qquad{\small --- Acc Drop}} & \cellcolor{mygray}{\color{blue}+2.66}  & \cellcolor{mygray}{\color{blue}+2.62}& \cellcolor{mygray}{\color{red}-3.30} & \cellcolor{mygray}{\color{blue}+2.08}  & \cellcolor{mygray}{\color{red}-1.63} & \cellcolor{mygray}{\color{red}-0.87} & \cellcolor{mygray}{\color{red}-0.46} & \cellcolor{mygray}{\color{red}-20.19} & \cellcolor{mygray}{\color{red}-6.60} & \cellcolor{mygray}{\color{red}-9.41} & \cellcolor{mygray}{\color{red}-3.51} \\

\rule{0pt}{2.2ex} & HiF4 &84.98
&87.95&71.19&84.04 &91.19&79.56&95.75&58.03&82.60&82.74&81.80\\

& \cellcolor{mygray}{\color{gray}\qquad{\small --- Acc Drop}} & \cellcolor{mygray}{\color{blue}+0.60}  & \cellcolor{mygray}{\color{blue}+1.31}& \cellcolor{mygray}{\color{blue}+4.34} & \cellcolor{mygray}{\color{blue}+1.95}  & \cellcolor{mygray}{\color{red}-0.27} & \cellcolor{mygray}{\color{red}-0.71} & \cellcolor{mygray}{\color{red}-0.16} & \cellcolor{mygray}{\color{red}-1.16} & \cellcolor{mygray}{\color{red}-2.20} & \cellcolor{mygray}{\color{blue}+1.09} & \cellcolor{mygray}{\color{blue}+0.48}\\

\hline 
\hline
\end{tabular}
\end{table*}

\noindent\textbf{Datasets:} 
We use a suite of widely adopted LLM benchmarks, 
including ARC-C(hallenge), ARC-E(asy)~\cite{Clark2018}, 
BoolQ~\cite{Clark2019}, HellaS(wag)~\cite{Zellers2019}, 
Lam(bada)Op(enAI)~\cite{Paperno2016}, 
Piqa~\cite{Bisk2020}, WinoG(rande)~\cite{Sakaguchi2019}, 
and MMLU~\cite{Hendrycks2021}. 
Together, these benchmarks span a broad spectrum of tasks, 
including commonsense and physical reasoning, 
cloze-style completion, pronoun/coreference resolution, 
fact verification, 
and multi-domain knowledge assessment. 

\noindent\textbf{Implementation Details:} 
Experiments in this section are all conducted using simulated 4-bit BFP 
formats based on Nvidia GPU and Huawei Ascend NPU. 
Specifically, before matrix multiplication, all linear layer tensors 
(with the exception of the embedding and LLM head layers) were converted 
from high-precision formats only to those that could be represented in 
HiF4 and NVFP4 formats. 
All experiments reported in this sub-section were conducted in parallel 
on two devices. 
For each device, we performed three runs using different random seeds. 
The final results are the average inference accuracy across the three runs 
and both devices to ensure that random variance is minimized. 
For a fair comparison, we evaluate all models and quantization settings 
within a unified framework, in which all experiments on a given benchmark 
employ the same prompt template for LLM inference. 

\noindent\textbf{Experimental Results and Analysis:} 
As shown in Tables~\ref{tab3.2.1_llama} and \ref{tab3.2.3_average}, 
HiF4 (direct-cast) consistently outperforms the prior state-of-the-art 
NVFP4 (direct-cast) and NVFP4+PTS across all four evaluated LLM models,  
demonstrating that the proposed HiF4 constitutes a more accurate 4-bit 
BFP format. 
We further observe that HiF4 is substantially more robust than NVFP4 
under direct conversion, 
attributable to its larger global and suitable local dynamic ranges. 
For instance, Mistral-7B exhibits a broader numerical distribution, 
under which direct-cast inference with NVFP4 results in model failure 
and accuracy degradation to near random-guess levels, 
\textit{i.e.}, inference crash. 
In contrast, HiF4 enables stable direct-cast inference under the same 
conditions. 
In addition, HiGPTQ further improves the inference 
accuracy of HiF4 across all four evaluated models. 
Notably, for the Qwen2.5-14B model, whose numerical distributions 
are optimized during training, 
the combination of HiF4 and HiGPTQ even surpasses the BF16 baseline. 

In summary, the proposed HiF4 format demonstrates higher inherent 
precision and stability than NVFP4. 
Many mature PTQ methods, such as GPTQ, can be adapted with minor 
modifications to further enhance the accuracy of HiF4.

\subsection{Experiments on DeepSeek-V3.1 and LongCat} 

\noindent\textbf{Models:} 
We further evaluate 4-bit inference 
accuracy on larger LLMs, DeepSeek-V3.1 (671B)~\cite{Liu2024} 
and LongCat (560B)~\cite{Team2025}. 
These two models further encompass Multi-Head Latent Attention 
(MLA) and Mixture-of-Experts (MoE) architectures, 
which are commonly used in latest state-of-the-art LLMs. 

\noindent\textbf{Datasets:} 
In addition to the benchmarks mentioned above, 
our evaluation suite extends to several high-stakes domains, 
specifically Gsm8K~\cite{Cobbe2021}, 
Math500~\cite{Hendrycks2021a} 
and CMMLU~\cite{Li2024a} 
to ensure a holistic assessment of the model's capabilities. 

\noindent\textbf{Implementation Details:} 
We deployed the LongCat and DeepSeek-V3.1 models using the vLLM 
framework on clusters of 32 and 64 Ascend 910B NPUs, 
respectively, distributed across 4 and 8 nodes. 
To ensure a standardized accuracy assessment, 
the AISBench framework was employed for evaluation. 
All experiments were completed within a practical timeframe. 
For 4-bit evaluation, we converted the tensors in the MLA\_linear, 
MoE\_linear (excluding the gating network), and FFN\_linear layers 
to HiF4 and NVFP4 formats. 

\noindent\textbf{Experimental Results and Analysis:} 
Results in Table~\ref{tab3.2.1_deepseek} demonstrate that 
HiF4 (direct-cast) consistently surpasses the state-of-the-art 
NVFP4 format in 4-bit inference accuracy, 
including NVFP4 (direct-cast) and NVFP4+PTS. 
The superiority of HiF4 stems from its novel structure design 
as depicted in Figure~\ref{HiF4-Structure}, 
and more balanced features as outlined in Table~\ref{HiF4-NVFP4}, 
which provide greater robustness for deep learning. 
For instance, HiF4 successfully handles the relatively  
quantization-sensitive LLM LongCat, where NVFP4 fails and leads 
to accuracy crashes in some hard tasks.  
Furthermore, in specialized cases like the DeepSeek-V3.1 model, 
the inherent design of HiF4 allows it to rival the BF16 
baseline, highlighting its effectiveness as a high-fidelity 
4-bit BFP format.

\section{Conclusion}

In this paper, we proposed a novel 4-bit BFP format, HiF4, for deep 
learning. 
Behind the design of HiF4, we first argue that E1M2 (S1P2) with 3-bit 
significand offers higher precision ceiling than E2M1 with 2-bit 
significand in the 4-bit BFP format design. 
Then, we successfully identified that increasing group size 
to 64 and employing two-level shared micro-exponents with a metadata 
overhead of 0.5-bit per value offers a new trade-off among 
overall accuracy, quantization complexity, and hardware efficiency. 
Additionally, we meticulously designed a dedicated global base scale 
that delivers sufficient inter-group dynamic range capable of directly 
supporting both training and inference scenarios, 
while enhancing the utilization of available intra-group dynamic range. 

Numerical analysis shows that, compared to the state-of-the-art 
NVFP4 format, HiF4 reduces the MSE quantization error by 24\% on  
Gaussian-distributed data. 
When 64-length dot product is integrated into existing hardware 
dot-product units originally optimized for 16-bit (FP16/BF16) 
and 8-bit (INT8/Float8) precision, 
HiF4 occupies only approximately one-third of the incremental 
area of NVFP4 and reduces power consumption by about 10\%. 
Finally, we conducted extensive LLM inference experiments, 
demonstrating that HiF4 consistently achieves higher accuracy 
than NVFP4 across various LLM architectures and diverse tasks. 
Future work will present the training potential of the proposed 
HiF4 format.


\bibliographystyle{IEEEtran}
\bibliography{IEEEabrv}

@InProceedings{Hendrycks2021a,
  author    = {Hendrycks, Dan and Burns, Collin and Kadavath, Saurav and Arora, Akul and Basart, Steven and Tang, Eric and Song, Dawn and Steinhardt, Jacob},
  booktitle = {NeurIPS},
  title     = {Measuring Mathematical Problem Solving With the MATH Dataset},
  year      = {2021},
}

@Article{Cobbe2021,
  author  = {Cobbe, Karl and Kosaraju, Vineet and Bavarian, Mohammad and Chen, Mark and Jun, Heewoo and Kaiser, Lukasz and Plappert, Matthias and Tworek, Jerry and Hilton, Jacob and Nakano, Reiichiro and Hesse, Christopher and Schulman, John},
  journal = {arXiv preprint arXiv:2110.14168},
  title   = {Training Verifiers to Solve Math Word Problems},
  year    = {2021},
}

@Article{Naveed2025,
  author    = {Naveed, Humza and Khan, Asad Ullah and Qiu, Shi and Saqib, Muhammad and Anwar, Saeed and Usman, Muhammad and Akhtar, Naveed and Barnes, Nick and Mian, Ajmal},
  journal   = {ACM Transactions on Intelligent Systems and Technology},
  title     = {A comprehensive overview of large language models},
  year      = {2025},
  number    = {5},
  pages     = {1--72},
  volume    = {16},
  publisher = {ACM New York, NY},
}

@Article{Gholami2024,
  author    = {Gholami, Amir and Yao, Zhewei and Kim, Sehoon and Hooper, Coleman and Mahoney, Michael W and Keutzer, Kurt},
  journal   = {IEEE Micro},
  title     = {AI and Memory Wall},
  year      = {2024},
  number    = {3},
  pages     = {33--39},
  volume    = {44},
  publisher = {IEEE},
}

@Article{AMD2025,
  author  = {AMD},
  journal = {https://www.amd.com/content/dam/amd/en/documents/instinct-tech-docs/product-briefs/amd-instinct-mi350x-gpu-brochure.pdf},
  title   = {AMD INSTINCT MI350X GPU Brochure},
  year    = {2025},
}

@Article{Luo2024,
  author  = {Luo, Yuanyong and Zhang, Zhongxing and Wu, Richard and Liu, Hu and Jin, Ying and Zheng, Kai and Wang, Minmin and He, Zhanying and Hu, Guipeng and Chen, Luyao and others},
  journal = {arXiv preprint arXiv:2409.16626},
  title   = {Ascend HiFloat8 Format for Deep Learning},
  year    = {2024},
}

@Article{Vaswani2017,
  author  = {Vaswani, Ashish and Shazeer, Noam and Parmar, Niki and Uszkoreit, Jakob and Jones, Llion and Gomez, Aidan N and Kaiser, {\L}ukasz and Polosukhin, Illia},
  journal = {Advances in neural information processing systems},
  title   = {Attention is all you need},
  year    = {2017},
  volume  = {30},
}

@Article{Micikevicius2025,
  author  = {Paulius Micikevicius},
  journal = {https://www.nvidia.com/en-us/on-demand/session/gtc25-s72458/},
  title   = {Blackwell Numerics for AI},
  year    = {2025},
}

@InProceedings{Clark2019,
  author    = {Christopher Clark and Kenton Lee and Ming-Wei Chang and Tom Kwiatkowski and Michael Collins and Kristina Toutanova},
  booktitle = {Proceedings of the 2019 Conference of the North American Chapter of the Association for Computational Linguistics: Human Language Technologies (NAACL-HLT ’19)},
  title     = {BoolQ: Exploring the Surprising Difficulty of Natural Yes/No Questions},
  year      = {2019},
  address   = {Minneapolis, MN, USA},
  pages     = {2924–2936},
  publisher = {Association for Computational Linguistics},
  doi       = {10.18653/v1/N19-1300},
}

@InProceedings{Liao2019,
  author       = {Liao, Heng and Tu, Jiajin and Xia, Jing and Zhou, Xiping},
  booktitle    = {2019 IEEE Hot Chips 31 Symposium (HCS)},
  title        = {DaVinci: A scalable architecture for neural network computing},
  year         = {2019},
  organization = {IEEE Computer Society},
  pages        = {1--44},
}

@Article{Liu2024,
  author  = {Liu, Aixin and Feng, Bei and Xue, Bing and Wang, Bingxuan and Wu, Bochao and Lu, Chengda and Zhao, Chenggang and Deng, Chengqi and Zhang, Chenyu and Ruan, Chong and others},
  journal = {arXiv preprint arXiv:2412.19437},
  title   = {DeepSeek-V3 Technical Report},
  year    = {2024},
}

@Article{Shazeer2020,
  author  = {Shazeer, Noam},
  journal = {arXiv preprint arXiv:2002.05202},
  title   = {Glu variants improve transformer},
  year    = {2020},
}

@Article{Agarwal2025,
  author  = {Agarwal, Sandhini and Ahmad, Lama and Ai, Jason and Altman, Sam and Applebaum, Andy and Arbus, Edwin and Arora, Rahul K and Bai, Yu and Baker, Bowen and Bao, Haiming and others},
  journal = {arXiv preprint arXiv:2508.10925},
  title   = {gpt-oss-120b \& gpt-oss-20b model card},
  year    = {2025},
}

@InProceedings{Frantar2023,
  author    = {Elias Frantar and Saleh Ashkboos and Torsten Hoefler and Dan Alistarh},
  booktitle = {Proceedings of the International Conference on Learning Representations (ICLR)},
  title     = {{GPTQ}: Accurate Post-training Compression for Generative Pretrained Transformers},
  year      = {2023},
}

@Article{Ainslie2023,
  author  = {Ainslie, Joshua and Lee-Thorp, James and De Jong, Michiel and Zemlyanskiy, Yury and Lebr{\'o}n, Federico and Sanghai, Sumit},
  journal = {arXiv preprint arXiv:2305.13245},
  title   = {Gqa: Training generalized multi-query transformer models from multi-head checkpoints},
  year    = {2023},
}

@Article{Eric2025,
  author  = {Xu Eric},
  journal = {Huawei Connect 2025, https://www.huawei.com/en/news/2025/9/hc-xu-keynote-speech},
  title   = {Groundbreaking SuperPoD Interconnect: Leading a New Paradigm for AI Infrastructure},
  year    = {2025},
}

@InProceedings{Zellers2019,
  author    = {Rowan Zellers and Ari Holtzman and Yonatan Bisk and Ali Farhadi and Yejin Choi},
  booktitle = {Proceedings of the 57th Annual Meeting of the Association for Computational Linguistics (ACL)},
  title     = {HellaSwag: Can a Machine Really Finish Your Sentence?},
  year      = {2019},
  address   = {Florence, Italy},
  pages     = {4791--4800},
  publisher = {Association for Computational Linguistics},
  doi       = {10.18653/v1/P19-1472},
}

@Article{Alvarez2025,
  author  = {Alvarez, Eduardo and Almog, Omri and Chung, Eric and Layton, Simon and Stosic, Dusan and Krashinsky, Ronny and Aubrey, Kyle},
  journal = {https://developer.nvidia.com/blog/introducing-nvfp4-for-efficient-and-accurate-low-precision-inference/},
  title   = {Introducing NVFP4 for Efficient and Accurate Low-Precision Inference},
  year    = {2025},
}

@Article{Touvron2023,
  author  = {Touvron, Hugo and Martin, Louis and Stone, Kevin and Albert, Peter and Almahairi, Amjad and Babaei, Yasmine and Bashlykov, Nikolay and Batra, Soumya and Bhargava, Prajjwal and Bhosale, Shruti and others},
  journal = {arXiv preprint arXiv:2307.09288},
  title   = {Llama 2: Open foundation and fine-tuned chat models},
  year    = {2023},
}

@Article{Team2025,
  author  = {Team, Meituan LongCat and Li, Bei and Lei, Bingye and Wang, Bo and Rong, Bolin and Wang, Chao and Zhang, Chao and Gao, Chen and Zhang, Chen and Sun, Cheng and others},
  journal = {arXiv preprint arXiv:2509.01322},
  title   = {LongCat-Flash Technical Report},
  year    = {2025},
}

@InProceedings{Hendrycks2021,
  author    = {Dan Hendrycks and Collin Burns and Steven Basart and Andy Zou and Mantas Mazeika and Dawn Song and Jacob Steinhardt},
  booktitle = {Proceedings of the International Conference on Learning Representations (ICLR) 2021},
  title     = {Measuring Massive Multitask Language Understanding},
  year      = {2021},
}

@Article{Rouhani2023,
  author  = {Rouhani, Bita Darvish and Zhao, Ritchie and More, Ankit and Hall, Mathew and Khodamoradi, Alireza and Deng, Summer and Choudhary, Dhruv and Cornea, Marius and Dellinger, Eric and Denolf, Kristof and others},
  journal = {arXiv preprint arXiv:2310.10537},
  title   = {Microscaling data formats for deep learning},
  year    = {2023},
}

@Misc{Jiang2023,
  author        = {Albert Q. Jiang and Alexandre Sablayrolles and Arthur Mensch and Chris Bamford and others},
  title         = {Mistral 7B},
  year          = {2023},
  archiveprefix = {arXiv},
  eprint        = {2310.06825},
  primaryclass  = {cs.CL},
  url           = {https://arxiv.org/abs/2310.06825},
}

@Article{Nvidia2025,
  author  = {Nvidia},
  journal = {https://resources.nvidia.com/en-us-blackwell-architecture/blackwell-architecture-technical-brief},
  title   = {NVIDIA Blackwell Architecture Technical Brief},
  year    = {2025},
}

@Article{Nvidia2022,
  author  = {Nvidia},
  journal = {https://resources.nvidia.com/en-us-hopper-architecture/nvidia-h100-tensor-c},
  title   = {NVIDIA H100 Tensor Core GPU Architecture},
  year    = {2022},
}

@Article{Nvidia2017,
  author  = {Nvidia},
  journal = {https://images.nvidia.com/content/volta-architecture/pdf/volta-architecture-whitepaper.pdf},
  title   = {NVIDIA TESLA V100 GPU ARCHITECTURE},
  year    = {2017},
}

@Article{Rouhani2023a,
  author  = {Rouhani, Bita Darvish and Garegrat, Nitin and Savell, Tom and More, Ankit and Han, Kyung-Nam and Zhao, Ritchie and Hall, Mathew and others},
  journal = {Open Compute Project},
  title   = {OCP Microscaling (MX) Specification},
  year    = {2023},
}

@InProceedings{Bisk2020,
  author    = {Yonatan Bisk and Rowan Zellers and Ronan Le Bras and Jianfeng Gao and Yejin Choi},
  booktitle = {Thirty-Fourth AAAI Conference on Artificial Intelligence (AAAI ’20)},
  title     = {PIQA: Reasoning about Physical Commonsense in Natural Language},
  year      = {2020},
  pages     = {7432--7439},
  doi       = {10.1609/AAAI.V34I05.6239},
}

@Article{Abecassis2025,
  author  = {Abecassis, Felix and Agrusa, Anjulie and Ahn, Dong and Alben, Jonah and Alborghetti, Stefania and Andersch, Michael and Arayandi, Sivakumar and Bjorlin, Alexis and Blakeman, Aaron and Briones, Evan and others},
  journal = {arXiv preprint arXiv:2509.25149},
  title   = {Pretraining Large Language Models with NVFP4},
  year    = {2025},
}

@Article{DarvishRouhani2020,
  author  = {Darvish Rouhani, Bita and Lo, Daniel and Zhao, Ritchie and Liu, Ming and Fowers, Jeremy and Ovtcharov, Kalin and Vinogradsky, Anna and Massengill, Sarah and Yang, Lita and Bittner, Ray and others},
  journal = {Advances in neural information processing systems},
  title   = {Pushing the limits of narrow precision inferencing at cloud scale with microsoft floating point},
  year    = {2020},
  pages   = {10271--10281},
  volume  = {33},
}

@Misc{Qwen2025,
  author        = {An Yang and Baosong Yang and Beichen Zhang and Binyuan Hui and Bo Zheng and Bowen Yu and Chengyuan Li and Dayiheng Liu and Fei Huang and others},
  title         = {Qwen2.5 Technical Report},
  year          = {2025},
  archiveprefix = {arXiv},
  eprint        = {2412.15115},
  primaryclass  = {cs.CL},
  url           = {https://arxiv.org/abs/2412.15115},
}

@InProceedings{Paperno2016,
  author    = {Paperno, Denis and Kruszewski, Germ{\'a}n and Lazaridou, Angeliki and Pham, Ngoc Quan and Bernardi, Raffaella and Pezzelle, Sandro and Baroni, Marco and Boleda, Gemma and Fern{\'a}ndez, Raquel},
  booktitle = {Proceedings of the 54th Annual Meeting of the Association for Computational Linguistics (Volume 1: Long Papers)},
  title     = {The {LAMBADA} dataset: Word prediction requiring a broad discourse context},
  year      = {2016},
  address   = {Berlin, Germany},
  pages     = {1525--1534},
  publisher = {Association for Computational Linguistics},
  doi       = {10.18653/v1/P16-1144},
}

@Article{Dubey2024,
  author  = {Dubey, Abhimanyu and Jauhri, Abhinav and Pandey, Abhinav and Kadian, Abhishek and Al-Dahle, Ahmad and Letman, Aiesha and Mathur, Akhil and Schelten, Alan and Yang, Amy and Fan, Angela and others},
  journal = {arXiv e-prints},
  title   = {The llama 3 herd of models},
  year    = {2024},
  pages   = {arXiv--2407},
}

@Article{Clark2018,
  author  = {Peter Clark and Isaac Cowhey and Oren Etzioni and Tushar Khot and Ashish Sabharwal and Carissa Schoenick and Oyvind Tafjord},
  journal = {arXiv:1803.05457},
  title   = {Think you have Solved Question Answering? Try ARC, the AI2 Reasoning Challenge},
  year    = {2018},
}

@Article{AMD2024,
  author  = {AMD},
  journal = {https://www.amd.com/content/dam/amd/en/documents/products/adaptive-socs-and-fpgas/versal/versal-ai-edge-gen2-product-brief.pdf},
  title   = {Versal AI Edge Series Gen2 Product Brief},
  year    = {2024},
}

@Article{Sakaguchi2019,
  author  = {Keisuke Sakaguchi and Ronan Le Bras and Chandra Bhagavatula and Yejin Choi},
  journal = {arXiv preprint arXiv:1907.10641},
  title   = {WinoGrande: An Adversarial Winograd Schema Challenge at Scale},
  year    = {2019},
}

@InProceedings{DarvishRouhani2023,
  author    = {Darvish Rouhani, Bita and Zhao, Ritchie and Elango, Venmugil and Shafipour, Rasoul and Hall, Mathew and Mesmakhosroshahi, Maral and More, Ankit and Melnick, Levi and Golub, Maximilian and Varatkar, Girish and others},
  booktitle = {Proceedings of the 50th Annual International Symposium on Computer Architecture},
  title     = {With shared microexponents, a little shifting goes a long way},
  year      = {2023},
  pages     = {1--13},
}

@Misc{Li2024a,
  author        = {Haonan Li and Yixuan Zhang and Fajri Koto and Yifei Yang and Hai Zhao and Yeyun Gong and Nan Duan and Timothy Baldwin},
  title         = {CMMLU: Measuring massive multitask language understanding in Chinese},
  year          = {2024},
  archiveprefix = {arXiv},
  eprint        = {2306.09212},
  primaryclass  = {cs.CL},
  url           = {https://arxiv.org/abs/2306.09212},
}

\end{document}